\documentclass[onecolumn]{IEEEtran}

\usepackage{amsfonts, amssymb, amsmath}
\usepackage{epsfig}
\usepackage{theorem}
\usepackage{textcomp}
\usepackage{multirow}
\usepackage{tabularx}
\usepackage{graphicx}
\usepackage{epstopdf}

\makeatletter
\let\NAT@parse\undefined
\makeatletter
\usepackage{natbib}

\newtheorem{theorem}{Theorem}

\newtheorem{lemma}[theorem]{Lemma}
\newtheorem{conjecture}[theorem]{Conjecture}

\newtheorem{problem}{Problem}
\newtheorem{objective}{Objective}

\def\remark{\noindent\textbf{Remark. }}

\newcommand{\qed}{\hfill $\Box$\\}
\newcommand{\rqed}{\hfill $\triangle$\\}

\def\mpp{\textrm{MPP}}

\def\mmpp{\textsc{M3PP}}
\def\tmpp{\textsc{MTATMPP}}
\def\dmpp{\textsc{MTDMPP}}
\def\mmdmpp{\textsc{MMDMPP}}

\def\sat{\textsc{3SAT}}
\def\satt{\textsc{2/2/4-SAT}}

\begin{document}

\title{Optimal Multi-Robot Path Planning on Graphs: Structure and Computational Complexity}
\author{
\begin{tabular}{ccc}
Jingjin Yu & & Steven M. LaValle 
\end{tabular}
\thanks{
Jingjin Yu is with the Computer Science and Artificial Intelligence Lab, Massachusetts Institute of Technology, Cambridge, MA 02139, USA. E-mail: jingjin@csail.mit.edu. Steven M. LaValle is with the Department of Computer Science, University of Illinois at Urbana-Champaign, Urbana, IL 61801 USA. E-mail: lavalle@illinois.edu.
}
}
\maketitle

\begin{abstract}We study the problem of optimal multi-robot path planning on graphs ($\mpp$) over four distinct minimization objectives: the total arrival time, the makespan (last arrival time), the total distance, and the maximum (single-robot traveled) distance. On the structure side, we show that each pair of these four objectives induces a Pareto front and cannot always be optimized simultaneously. Then, through reductions from $\sat$, we further establish that computation over each objective is an NP-hard task, providing evidence that solving $\mpp$ optimally is generally intractable. Nevertheless, in a related paper \cite{YuLav15TRO-II}, we design complete algorithms and efficient heuristics for optimizing all four objectives, capable of solving $\mpp$ optimally or near-optimally for hundreds of robots in challenging setups.
\end{abstract}

\section{Introduction}
\label{sec:intro}
In this paper, we study the problem of optimal {\em multi-robot path planning on graphs} ($\mpp$), focusing on {\em structural and computational complexity} issues. In an $\mpp$ instance, the robots are uniquely labeled and uniform sized spheres confined to an arbitrary connected graph. The robots may move from a vertex to an adjacent one in one time step in the absence of collision, which may occur when two robots simultaneously move to the same vertex or along the same edge in different directions. Over the basic $\mpp$ formulation, we look at four minimization objectives: the total arrival time, the makespan (last arrival time), the total distance, and the maximum (single-robot traveled) distance. In addition to showing that each pair of these objectives induces a Pareto optimal structure, we prove that all four objectives are NP-hard to optimize. 

{\bf Related work.} Research on discrete multi-robot path planning problems can be traced back to the mathematical study of the $15$-puzzle \cite{Loy59}, in which $15$ labeled ($1$-$15$) square game pieces, confined on a $4\times 4$ grid, must be moved to a row major ordering from some initial configuration. Note that only a single piece may be moved at a time. In \cite{Sto1879}, it is observed that the feasibility of the $15$-puzzle is decided by the {\em parity} of the game setup. A while later, Wilson generalizes the observation, proving that the feasibility of moving $n-1$ labeled {\em pebbles} on an $n$-vertex $2$-connected graph depends on whether the graph is bipartite \cite{Wil74}. An algorithm for solving a feasible instance is also supplied in the work. A further generalization to arbitrary connected graph and arbitrary number of pebbles (but less than $n$, the number of vertices of the underlying graph) is proposed in \cite{KorMilSpi84}, which also gives a bound of $\Theta(n^3)$ as the number of moves required for solving a feasible instance. 

Motivated by applications toward computer games and multi-robot systems, concurrent movements are introduced and studied extensively in the past decade \cite{Sil05,JanStu08,LunBer11,WagChoC11,StaKor11}. We refer to these problems under the umbrella term of {\em multi-robot path planning}, or $\mpp$.\footnote{Another common name for these problems is {\em coorperative path-finding}.} Although the study in \cite{KorMilSpi84} assumes a single pebble (agent/robot) movement per time step, the $\Theta(n^3)$ bound on the number of moves continues to hold when synchronous robot movements are allowed. With multi-robot applications in mind, in \cite{YuRus14WAFR}, the $\Theta(n^3)$ bound is shown to extend when there are as many robots as graph vertices. As finding feasible solutions is a largely solved issue, most multi-robot path planning work has an emphasis on optimality.

A lingering question in the pursuit of efficient algorithms for optimal $\mpp$ is the structure and complexity of such problems. In particular, if it is unclear whether an optimal $\mpp$ formulation is computationally intractable, then polynomial-time algorithm should be sought after. When there are $n-1$ robots on a $n$-vertex graph, finding the shortest sequence of moves that solves a given problem has long been established as NP-hard \cite{Gol84}, even when the underlying graph is a grid \cite{RatWar90}. Note that in such cases, because only a single move is allowed in a time step, time- and distance-based optimality objectives are equivalent. When there are less than $n - 1$ robots with synchronous robot movements allowed, it is unclear that time- and distance-based objectives are again the same. Moreover, results like \cite{Gol84, RatWar90} do not carry over to show that optimal $\mpp$ with synchronous moves are again intractable. The only available complexity result for optimal $\mpp$ can be found in \cite{Sur10}, which shows that minimum makespan $\mpp$ is intractable through a rather involved reduction. 

{\bf Contributions.} To resolve issues surrounding the structural and complexity of optimal $\mpp$, in this work, we carry out a systematic study on optimal $\mpp$ covering four global and arguably the most common time- and distance-based minimization objectives. First, for each pair of the four objectives, through an infinite class of problem instances, we show that the pair of objectives cannot be optimized by a single solution. Moreover, the difference\footnote{For a pair of objectives, let $(a_1^*, a_2)$ and $(a_1, a_2^*)$ be two solution vectors on the Pareto front such that $a_1^*$ is the optimal solution for the first objective and $a_2^*$ the optimal solution for the second objective. The difference may be measured as $\| (a_1^*, a_2) - (a_1, a_2^*) \|_{\infty}$ with $\|\cdot \|_{\infty}$ being the $L^{\infty}$-norm.} between optimal solutions for these objectives can be arbitrarily large. Second, through reductions from the classical NP-hard $\sat$ problem, we prove that optimizing each of the four objectives is computationally intractable. We further establish that some of these problems remain hard even when there are only two groups of interchangeable robots (optimal $\mpp$ is solvable in polynomial time when there is a single group of robots \cite{YuLav13STAR}). Our relatively straightforward reduction schemes preserve the structure of $\sat$, clearly illustrating the source of difficulty that renders optimal $\mpp$ problems hard. 

The rest of the paper is organized as follows. In Section \ref{sec:formulation}, we define the four optimal $\mpp$ problems and provide necessary background materials. We discuss the Pareto optimal structures in Section \ref{sec:pareto} and prove the NP-hardness of optimizing these objectives in Section~\ref{sec:time} and Section~\ref{sec:distance}. We conclude in Section \ref{sec:conclusion}. This paper is partly based on \cite{YuLav13AAAI}. In comparison to \cite{YuLav13AAAI}, an additional optimality objective is introduced. Most significantly, we have devised new and much simplified proofs for the distance-optimal cases. 

\section{Preliminaries}\label{sec:formulation}
We now define $\mpp$ and the optimality objectives studied in this paper. Following the problem statements, we provide a brief description of $\sat$, a version of the {\em boolean satisfiability problem}, on which our complexity proofs are based. 

\subsection{Multi-Robot Path Planning on Graphs}
\begin{figure}[htp]
\begin{center}
  \begin{tabular}{ccc}
    \includegraphics[width=0.12\textwidth]{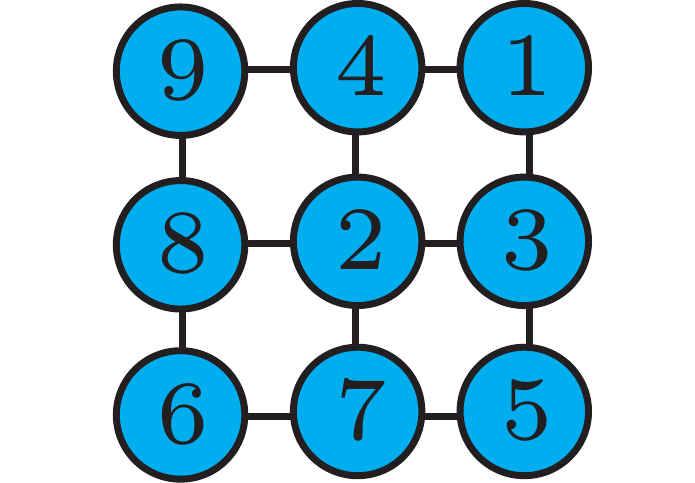} & &
    \includegraphics[width=0.12\textwidth]{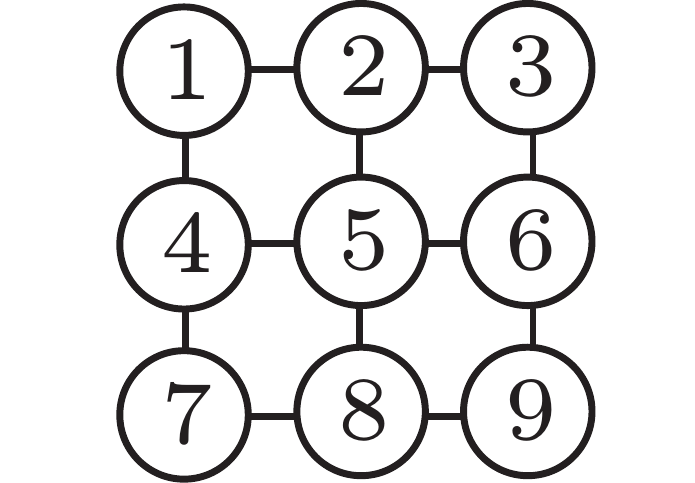}  \\
    (a) && (b)\\
  \end{tabular}
\end{center}
\caption{\label{fig:example} a) A 9-puzzle problem. b) The desired goal configuration.}
\end{figure}

Let $G = (V, E)$ be a connected, undirected, simple graph, with $V = \{v_i\}$ being the vertex set and $E = \{(v_i, v_j)\}$ the edge set. Let $R = \{r_1, \ldots, r_n\}$ be a set of $n$ robots. The robots move at discrete time steps ({\em i.e.}, at $t = 0, 1, \ldots$). At time step $t = 0$, each robot occupies a distinct vertex of $G$. In general, at any time step $t = 0, 1, \ldots$, the robots assume a {\em configuration} that is an injective map from $R$ to $V$. The {\em start} (or {\em initial}) and {\em goal} configurations of the robots are denoted as $x_I$ and $x_G$, respectively. As an example, Fig.~\ref{fig:example}(a) shows a possible configuration of $9$ robots on a $3\times 3$ grid graph.\footnote{In this paper, we generally use shaded discs to mark start locations of robots and discs without shading for goal locations.} Fig.~\ref{fig:example}(b) shows a possible goal configuration, in which the robots are ordered according to what is commonly known as the row major ordering.

\begin{figure}[ht!]
\begin{center}
  \begin{tabular}{ccc}
    \includegraphics[width=0.12\textwidth]{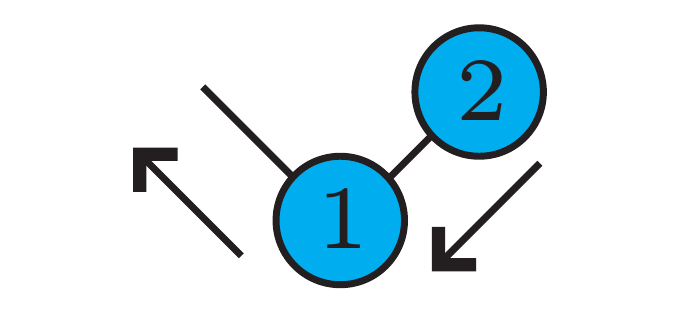} & 
    \includegraphics[width=0.09\textwidth]{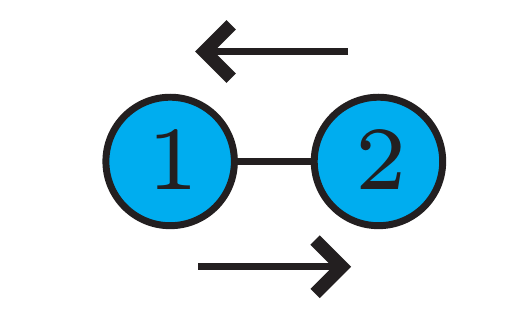} & 
    \includegraphics[width=0.09\textwidth]{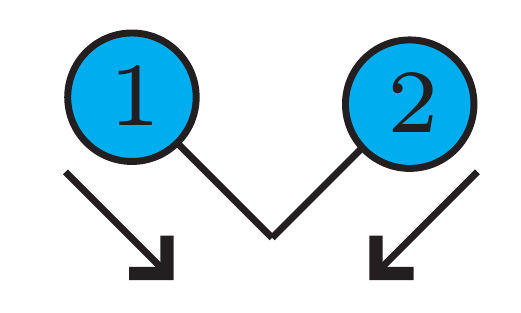}  \\
    (a) & (b) & (c)\\
  \end{tabular}
\end{center}
\caption{\label{fig:moves} Some feasible and infeasible moves for two robots. a) A feasible synchronous move. b) An infeasible synchronous move in which two robot collide ``head-on''. c) An infeasible synchronous move in which two robots ``meet'' at a vertex.}
\end{figure}

For robot motion, during a discrete time step, each robot may either stay at its current vertex or move to an adjacent vertex. To formally describe a plan, let a {\em scheduled path} be a map $p_i: \mathbb Z^+ \to V$, in which $\mathbb Z^+ := \mathbb N \cup \{0\}$. A scheduled path $p_i$ is {\em feasible} if it satisfies the following properties: 1) $p_i(0) = x_I(r_i)$. 2) For each $i$, there exists a smallest $t_i \in \mathbb Z^+$ such that $p_i(t_i) = x_G(r_i)$. 3) For any $t \ge t_i$, $p_i(t) \equiv x_G(r_i)$. 4) For any $0 \le t < t_i$, $(p_i(t), p_i(t+1)) \in E$ or $p_i(t) = p_i(t+1)$ (if $p_i(t) = p_i(t+1)$, robot $r_i$ stays at vertex $p_i(t)$ between the time steps $t$ and $t+1$). We say that two paths $p_i, p_{j}$ are in {\em collision} if there exists $k \in \mathbb Z^+$ such that $p_i(t) = p_{j}(t)$ (meet collision) or $(p_i(t), p_i(t+1)) = (p_j(t+1), p_j(t))$ (head-on collision). As an illustration, Fig.~\ref{fig:moves} shows the feasible and infeasible moves for two robots during a single time step \footnote{We assume that the graph $G$ only allows ``meet'' or ``head-on'' collisions. The assumption is mild. For example, a (arbitrary dimensional) grid with unit edge distance is such a graph for robots of with radii of no more than $\sqrt{2}/4$.}. The {\em multi-robot path planning on graph} (\mpp) problem is defined as follows. 

\begin{problem}[Multi-robot Path Planning on Graphs]\label{mpp} Given a $4$-tuple $(G, R, x_I, x_G)$, find a set of paths $P = \{p_1, \ldots, p_n\}$ such that $p_i$'s are feasible paths for respective robots $r_i$'s and no two paths $p_i, p_j$ are in collision. 
\end{problem}

For example, Fig.~\ref{fig:example}(a) and Fig.~\ref{fig:example}(b) define an $\mpp$ problem on the $3\times 3$ grid. We call this particular problem the {\em 9-puzzle} problem (a variant of the 15-puzzle \cite{RatWar90}), which readily generalizes to $N^2$-puzzles. 
\vspace{2mm}

\remark With a few exceptions ({\em e.g.}, \cite{StaKor11}), most existing studies on discrete multi-robot path planning problems require empty vertices as swap spaces. In contrast, our $\mpp$ formulation allows synchronized rotations of robots along fully occupied cycles (see, {\em e.g.}, Fig. \ref{fig:example} and \ref{fig:puzzle-8-sol}). 
~\rqed

\subsection{Optimal Formulations}
Let $P = \{p_1, \ldots, p_n\}$ be an arbitrary feasible solution to some fixed $\mpp$ instance. For a path $p_i \in P$, let $len(p_i)$ denote the length of the path $p_i$, which is increased by one each time when the robot $r_i$ passes an edge. A robot, following a path $p_i$, may visit the same edge multiple times. Recall that $t_i$ denotes the arrival time of robot $r_i$. In the study of optimal $\mpp$, we examine four most common, global objectives with two focusing on time optimality and two focusing on distance optimality. Below, each objective is defined formally, followed by the corresponding decision version of the $\mpp$ problem. Note that these decision versions are necessary for stating the complexity ({\em i.e.} NP-completeness) results.

\begin{objective}[Total Arrival Time]\label{ott} Compute a path set $P$ that minimizes 
$\sum_{i = 1}^nt_i$.
\end{objective}

\vspace{1mm}
\noindent $\tmpp$ (Minimum Total Arrival Time $\mpp$) \\
\noindent INSTANCE: An instance of $\mpp$, and $K \in \mathbb Z$.\\
\noindent QUESTION: Is there a solution path set $P$ with a total arrival time no more than $K$? 

\begin{objective}[Makespan]\label{omakespan} Compute a path set $P$ that minimizes 
$\max_{1 \le i \le n}t_i$.
\end{objective}

\vspace{1mm}
\noindent $\mmpp$ (Minimum Makespan $\mpp$) \\
\noindent INSTANCE: An instance of $\mpp$, and $K \in \mathbb Z$.\\
\noindent QUESTION: Is there a solution path set $P$ with a makespan no more than $K$? 

\begin{objective}[Total Distance]\label{otd}Compute a path set $P$ that minimizes 
$\sum_{i = 1}^n len(p_i)$.
\end{objective}

\vspace{1mm}
\noindent $\dmpp$ (Minimum Total Distance $\mpp$) \\
\noindent INSTANCE: An instance of $\mpp$, and $K \in \mathbb Z$.\\
\noindent QUESTION: Is there a solution path set $P$ with a total path distance no more than $K$? 

\begin{objective}[Maximum Distance]\label{md}Compute a path set $P$ that minimizes 
$\max_{1 \le i \le n}len(p_i)$.
\end{objective}

\vspace{1mm}
\noindent $\mmdmpp$ (Minimum Maximum Distance $\mpp$) \\
\noindent INSTANCE: An instance of $\mpp$, and $K \in \mathbb Z$.\\
\noindent QUESTION: Is there a solution path set $P$ in which every path has a distance no more than $K$? 
\vspace{1mm}

\begin{figure}[htp]
\begin{center}
    \includegraphics[width=0.48\textwidth]{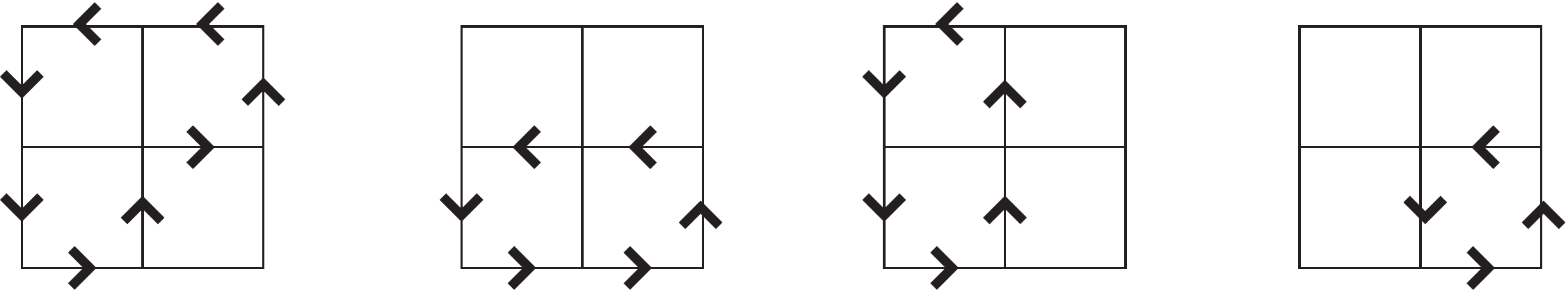} 
\end{center}
\caption{\label{fig:puzzle-8-sol} A 4-step solution from our algorithm. The directed edges show the moving directions of the robots at the tail of the edges.}
\end{figure}

The intuitive meaning of these objectives is clear from the definitions. Here, we provide a concrete example of a minimum makespan solution to the {\em 9-puzzle} problem given in Fig.~\ref{fig:example}. Since there is no empty vertex, the robots can only rotate together with other robots in a synchronous manner. A solution with minimum makespan is given in Fig. \ref{fig:puzzle-8-sol}. The time optimality of the solution is evident as it takes at least four steps for robot 9 to reach its goal. 

\subsection{The $\sat$ Problem}

All NP-hardness proofs in this paper are based on many-one reductions from NP-complete versions of the boolean satisfiability problem. In comparison to the conference paper \cite{YuLav13AAAI}, we have made an attempt to unify the proof strategy so that all reductions in the current work are based on the classical $\sat$ problem \cite{GarJoh79}. In addition to greatly simplifying the NP-hardness proof for distance optimal $\mpp$ problems (see Section~\ref{sec:distance}), the unification suggests that time- and distance-optimal $\mpp$ problems have the same source of complexity, {\em i.e.}, the intractability arises from the sharing of paths between robots traveling in opposite directions. 

A $\sat$ instance is defined by a 2-tuple $(X, C)$ in which $X = \{x_1, \ldots, x_n\}$ is the set of {\em binary variables} and $C = \{c_1, \ldots, c_m\}$ is the set of {\em disjunctive clauses}. Each clause $c_j \in C$ takes the form of $c_j = y_j^1\vee y_j^2 \vee y_j^3$, with each $y_j^k, 1 \le k \le 3$ a {\em literal}. That is, $y_j^k \in \{x_1, \neg x_1, \ldots, x_n, \neg x_n\}$. Without loss of generality, we may assume that for fixed $1 \le j \le m$, $y_j^1, y_j^2$, and $y_j^3$ are all distinct and do not contain literals of the same variable. An {\em assignment} of true or false values to the variables is represented as $\{\widetilde{x_1}, \ldots, \widetilde{x_n}\}$. Throughout this paper, the $\sat$ instance 

\begin{equation}\label{equation:sat}
\begin{array}{l}
X = \{x_1, x_2, x_3, x_4\}, \\ 
C = \{x_1\vee \neg x_3 \vee x_4, \neg x_1 \vee x_2 \vee \neg x_4, \neg x_2 \vee x_3 \vee x_4\}
\end{array}
\end{equation}
is employed when a concrete $\sat$ example is required. 

\section{Pareto Optimal Structure Between Objectives}\label{sec:pareto}
When it comes to optimization problems with multiple objectives, one generally expects a Pareto optimal structure. That is, fixing a problem instance, it is often unlikely that the optimal solution for one objective is also the optimal solution for a second objective. Optimal $\mpp$ is no exception: the solution vectors for each pair of Objectives~\ref{ott} to~\ref{md} form a Pareto front. In this section, for each pair of objectives, we provide an infinite family of $\mpp$ instances for which the two objectives are optimized by different solutions. This structural study promotes our understanding of the target problem, which in turn helps us design algorithms for solving the problem. We start with the two time-based objectives. 

\begin{lemma}\label{l:pareto-1}For $\mpp$, optimality cannot always be simultaneously achieved for minimum total arrival time and minimum makespan. 
\end{lemma}
\begin{figure}[htp]
\begin{center}
    \includegraphics[width=0.30\textwidth]{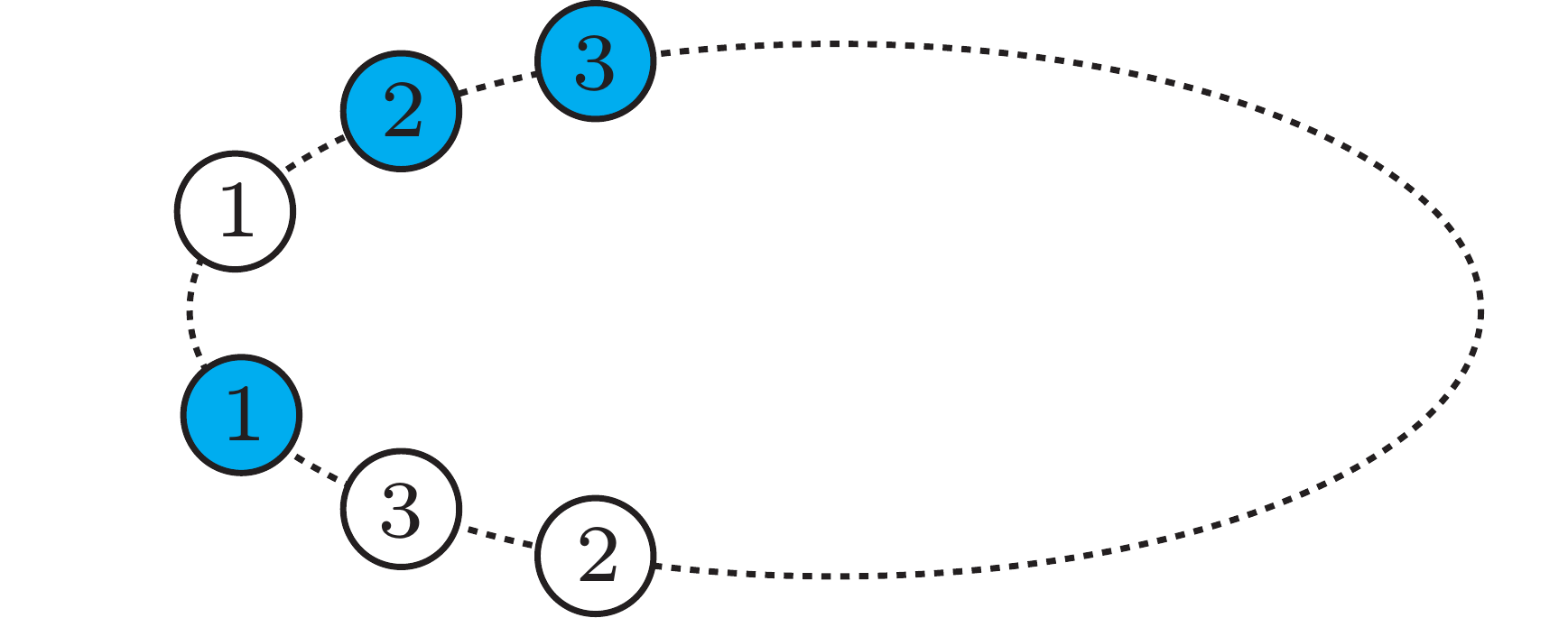} 
\end{center}
\caption{\label{fig:ms-tt} A class of problems for which minimum makespan and minimum total arrival time cannot be simultaneously achieved in a single solution.}
\end{figure}
{\sc Proof.} In Fig. \ref{fig:ms-tt}, the start and goal vertices of robots $1$-$3$ are as marked. Let the distance between each pair of the consecutive discs on the left side of the oval be one. Let the distance of the longer path on the right (between robot 3's starting location and 2's goal location) be some $x \ge 1$. Given the arrangement of the robots, optimal solutions require the robots to all move in the clockwise direction or all move in the counterclockwise direction until they reach their goals. If the robots all move in the clockwise direction, the cost vector for total arrival time and makespan is $(2x + 3, x + 1)$. The cost vector becomes $(x + 12, x + 4)$ when the robots move in the counterclockwise direction. Thus, clockwise moves always yield solutions with minimum makespans. However, when $x > 9$, the solution corresponding to counterclockwise movements has a smaller total arrival time. 
~\qed

The construction from Fig.~\ref{fig:ms-tt} also applies to show the general incompatibility between sum ({\em i.e.}, total distance or time) objectives and maximum ({\em i.e.}, maximum distance and time) objectives. 

\begin{lemma}\label{l:pareto-1x}For $\mpp$, optimality cannot always be simultaneously achieved for (i) minimum total distance and minimum maximum distance, (ii) minimum total time and minimum maximum distance, and (iii) minimum total distance and minimum makespan.  
\end{lemma}
{\sc Proof.} The solution vectors from the proof of Lemma~\ref{l:pareto-1} coincide with the solution vectors for maximum distance and total distance; the clockwise and counterclockwise solutions yield solution vectors with total distance and maximum distance as $(2x + 3, x + 1)$ and $(x + 12, x +4)$, respectively. Therefore, total distance and maximum distance objectives cannot be simultaneously minimized, which proves the claim for {\em (i)}. Following the same reasoning, the claim also holds for {\em (ii)} and {\em (iii)}. ~\qed

For the other two pairings, a different set of problems are needed.  

\begin{lemma}\label{l:pareto-3}For $\mpp$, optimality cannot always be simultaneously achieved for minimum total arrival time and minimum total distance. 
\end{lemma}
\begin{figure}[htp]
\begin{center}
    \includegraphics[width=0.30\textwidth]{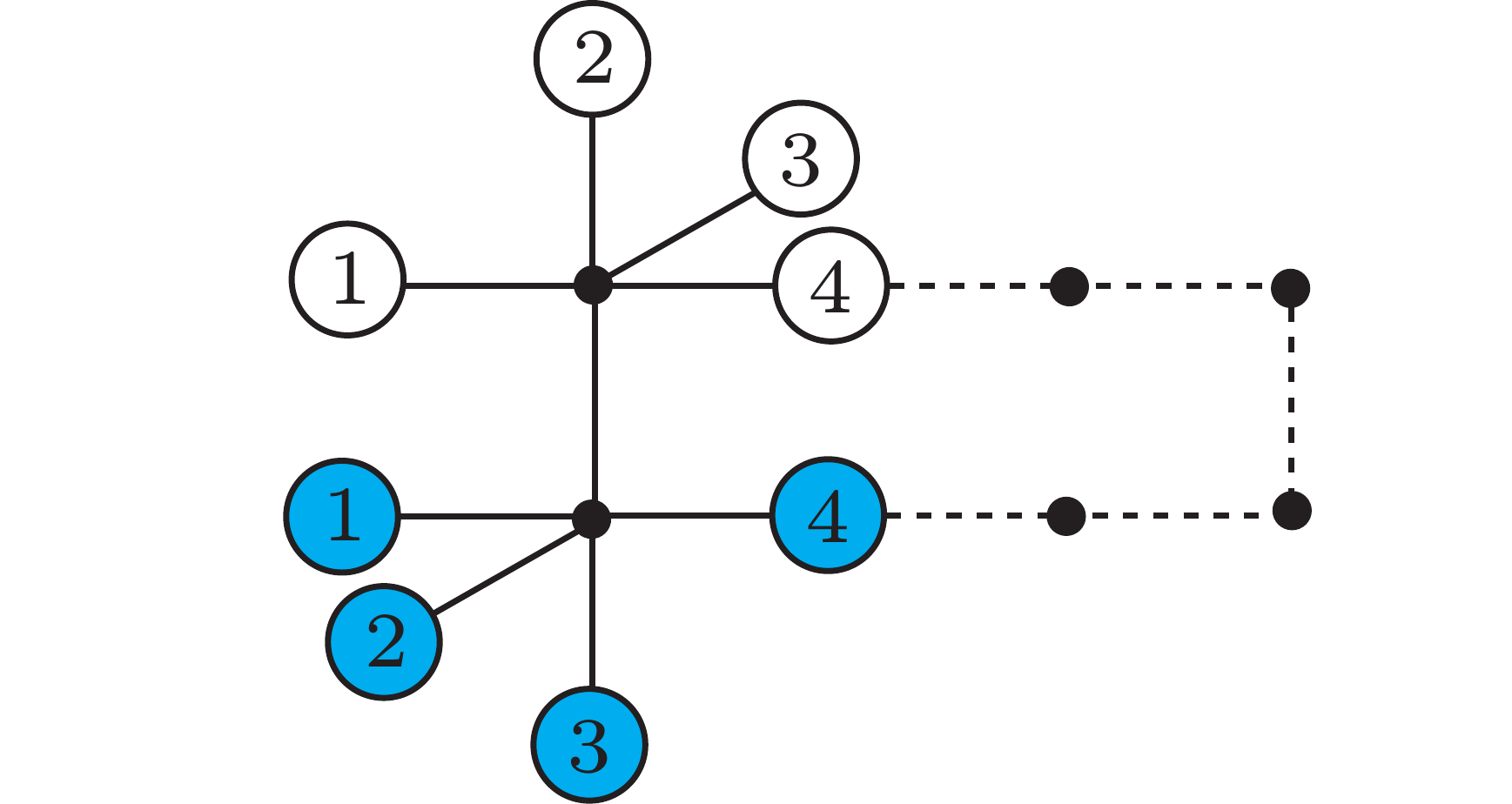} 
\end{center}
\caption{\label{fig:tt-td} A class of problems for which minimum total arrival time and minimum total distance cannot be simultaneously achieved in a single solution.}
\end{figure}
{\sc Proof.} In Fig. \ref{fig:tt-td}, the starts and goals of robots $1$-$4$  are as marked. The distance between any adjacent pair of nodes (discs and black dots) is one. The solution with minimum total arrival time sends robots $1$-$3$ through the solid path on the left and robot 4 through the dotted path on the right. This yields a total arrival time of $3 + 4 + 5 + 4 = 16$ and a total distance of $3 + 3 + 3 + 4 = 13$. On the other hand, the solution with minimum total distance sends all robots from the left path, which yields a total arrival time of $18$ and a total distance of $12$. By extending the lengths of the two vertical edges in the middle, we get an infinite family of examples. ~\qed

\remark Using the construction from Fig.~\ref{fig:tt-td}, we could show that maximum distance of an $\mpp$ solution cannot always be minimized in conjunction with makespan. To see this, note that the solution that minimize makespan requires robot $4$ to go through the right vertical edge, which inevitably lengthens the distance traveled by the robot. ~\rqed

With Lemmas~\ref{l:pareto-1}-\ref{l:pareto-3} and the accompanying remark, we reach the following conclusion concerning the structures of optimal $\mpp$ problems. 

\begin{theorem}\label{t:pareto}For $\mpp$, a solution that simultaneously minimizes any pair of objectives from Objectives~\ref{ott}-\ref{md} cannot always be found. \end{theorem}


\section{Computational Complexity of Time-Optimal $\mpp$ Formulations}\label{sec:time}
In this section and Section~\ref{sec:distance}, we show that optimizing each of the Objectives \ref{ott}-\ref{md} is NP-hard. We devote this section to the discussion of time optimality. First, we prove that computing a minimum total arrival time solution is an intractable task. 

\begin{figure}[ht]
\begin{center}
\includegraphics[width=0.48\textwidth]{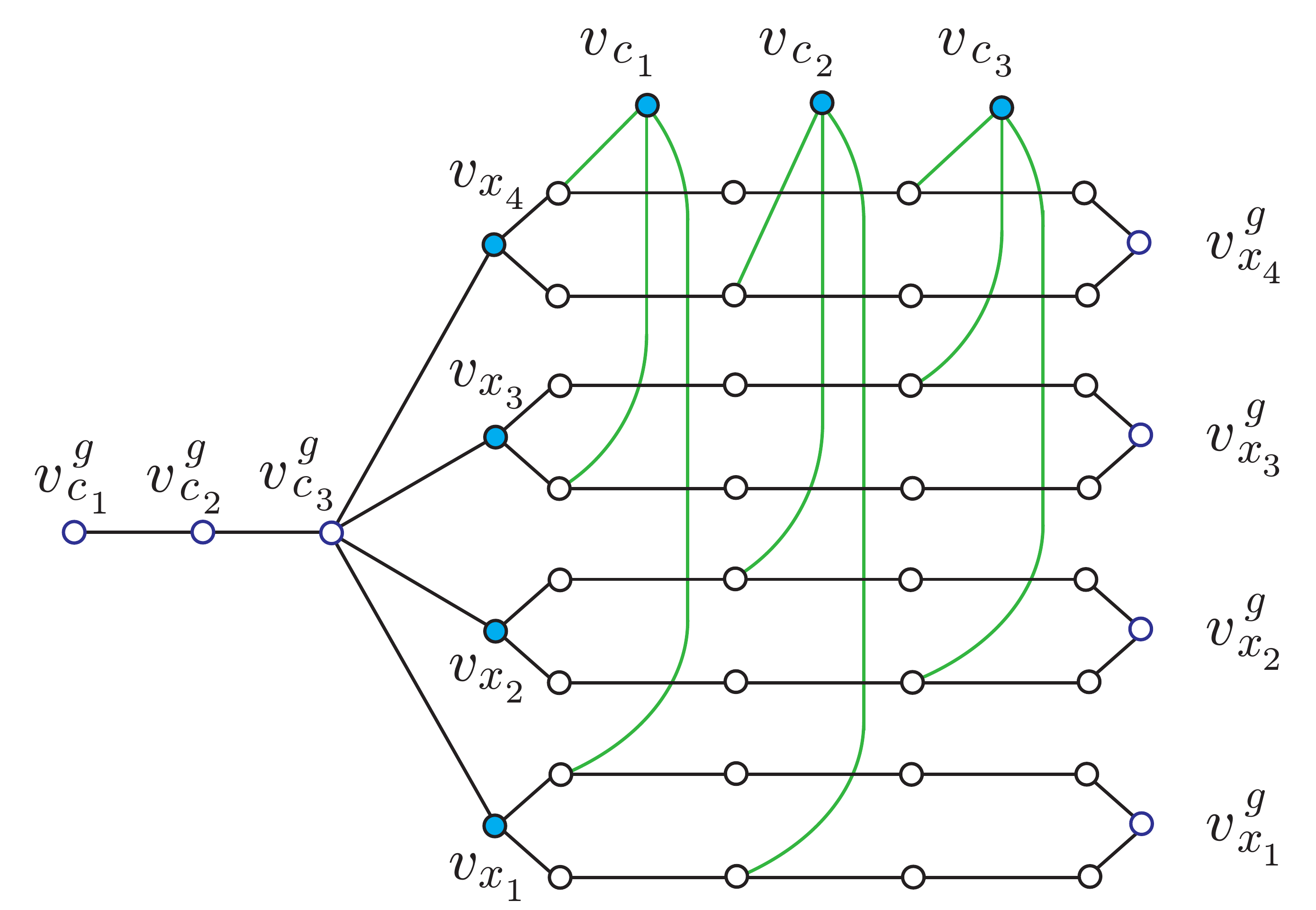} 
\end{center}
\caption{\label{fig:3sat-mpp} Reduction of $\sat$ to $\tmpp$ with $K = 5$. }
\end{figure}

\begin{theorem}\label{t:mpp-tt-np-hard} $\tmpp$ is NP-complete.
\end{theorem}
\noindent{\sc Proof.} {\bf Instance construction.} We reduce from $\sat$ to $\tmpp$. From a $\sat$ instance (the $\sat$ instance~\eqref{equation:sat} is used as the example), an $\tmpp$ instance is constructed as follows. For each variable $x_i$, two paths of length $m + 2$ each, joined at the end, are added. For our example, this step yields the four horizontal strips, stacked vertically, in the middle of Fig. \ref{fig:3sat-mpp}. For convenience, we call the two paths of the $i$-th horizontal strip (for variable $x_i$) the $i$-th upper and lower paths. At the left end of the construct ({\em i.e.}, vertex $v_{x_i}$) sits a {\em variable robot} $r_{x_i}$, with its goal vertex $v_{x_i}^g$ at the right end. Robot $r_{x_i}$ needs at least $m + 2$ steps to reach it goal. 

Next, for each clause $c_j = y_j^1 \vee y_j^2 \vee y_j^3$, a {\em clause} robot $r_{c_j}$ is set to start from the vertex $v_{c_j}$ (see Fig. \ref{fig:3sat-mpp}). The vertex $v_{c_j}$ is connected to three paths associated with the three variables corresponding to $c_j$'s three literals. If a literal takes the non-negated (resp., negated) form of variable $x_i$, then $v_{c_j}$ is connected to the $i$-th upper (resp., lower) path at a vertex of distance $j$ from $v_{x_i}$. For example, for $c_1 = x_1 \vee \neg x_3 \vee x_4$, $v_{c_1}$ is connected to the first upper, third lower, and fourth upper paths, all at vertices of distance $1$ from the left ends of the horizontal strips. 

After the clause structures are created, the goals for the $r_{c_j}$'s are added. For this purpose, a path of length $m - 1$ is added ({\em e.g.} the leftmost path with blue vertices in Fig. \ref{fig:3sat-mpp}), with the left vertex being the goal for $r_{c_1}$ and the right vertex the goal for $r_{c_m}$. The goal vertex for $r_{c_m}$, $v_{c_m}^g$, is connected to all $v_{x_i}$'s, the start vertices of robots $r_{x_i}$'s. Having constructed an $\mpp$ instance, setting $K = (n + m)(m + 2)$ fully describes an instance of $\tmpp$. Fig. \ref{fig:3sat-mpp} gives the complete graph for the $\tmpp$ instance constructed from the example $\sat$ instance. Here, $n = 4$ and $m = 3$, which makes $K = 35$. 

{\bf Many-one reduction}. If the $\sat$ instance is satisfiable, let $\widetilde{x_1}, \ldots, \widetilde{x_n}$ be an assignment of the truth values to the variables. For each variable $x_i$, if $\widetilde{x_i}$ is true (resp., false), then let robot $r_{x_i}$ take the lower (resp., upper) path on its strip. The upper (resp., lower) path is then free to use for transporting the robots corresponding to the clauses, $r_{c_j}$'s. All $m + n$ robots can start moving at time step zero and arrive at their desired goals at time step $m + 2$. The total time is then $(m + n)(m+2)$. 

On the other hand, if the $\mpp$ instance has a solution with total arrival time $(n + m)(m + 2)$, then every robot must start moving at time step zero, follow a shortest path, and never stop until it reaches its goal. This forces every robot $r_{x_i}$ to take either the upper or lower path on its own horizontal strip, which prevents any robot $r_{c_j}$ from using the same path in the opposite direction. If robot $r_{x_i}$ uses the upper (resp., lower) path, let $\widetilde{x_i} = false$ (resp., $true$). The resulting assignment $\widetilde{x_1}, \ldots, \widetilde{x_n}$ satisfies the $\sat$ instance. This proves the NP-hardness of $\tmpp$. Because $\tmpp$ is also in NP (checking that a given solution uses no more than some total time $K$ is easy), it is NP-complete. ~\qed

\remark $\sat$'s complexity arises because the values of many binary variables in an instance, which model input signals to a set of connected logical gates, must be simultaneously decided. In the proof of Theorem~\ref{t:mpp-tt-np-hard}, we could simulate the logical-gate-like structure of $\sat$ due to the time synchronization among the robots required from an optimal solution. Since all robots must move the same number of time steps in our construction, the number of candidate paths is very limited. There are two such paths for each variable robot and three such paths for a clause robot. Then, the individual paths for the clause robots effectively simulate a integrated circuit. The choice among three possible paths simulates an {\em or} gate and sequentially arranged goals simulate an {\em and} gate. The forced time synchronization also enabled the use a minimum number of robots; only $n$ (the number of variable) robots are used to simulate the choice between true and false for a variable, and only $m$ robots are used to simulate the conjunctions of $m$ disjunctive clauses.~\rqed 

The carefully constructed reduction scheme for proving Theorem~\ref{t:mpp-tt-np-hard} readily extends to show that minimizing the makespan is also NP-hard. 

\begin{theorem}\label{t:mpp-np-hard} $\mmpp$ is NP-complete.
\end{theorem}
{\sc Proof.} In the proof of Theorem \ref{t:mpp-tt-np-hard}, after the $\mpp$ instance is created, setting $K = m + 2$ as the minimum makespan produces an $\mmpp$ instance from the $\sat$ instance. The rest of the proof remains the same. ~\qed

\remark Previously, complexity results on optimal $\mpp$ problems mostly focus on variants of the 15-puzzle with a single empty vertex for swapping. In that case, only one robot may move in a time step, thus rendering time optimality equivalent to distance optimality. This is no longer the case here. One exception is \cite{Sur10}, which addresses the time optimality of parallel pebble motion problems. As explained in \cite{YuRus14WAFR}, Theorem \ref{t:mpp-np-hard} can be established through a reduction from the NP-hard minimum makespan {\em parallel pebble motion} problem from \cite{Sur10}. The proof technique given in \cite{Sur10} is however highly involved and seems unnecessarily complex. The proof presented here, in addition to working for proving Theorem \ref{t:mpp-tt-np-hard}, is much more direct. Moreover, the reduction ({\em e.g.}, Fig. \ref{fig:3sat-mpp}) clearly illustrates what makes finding time optimal solutions hard: When multiple robots move in opposite directions thorough a few shared paths, it is critical that the right paths are picked if time optimality is desired. 

In fact, our proof technique allows us to establish an even stronger intractability result that is surprising: computing a time optimal solution is NP-hard even when there are only two groups of interchangeable robots. By a {\em group of interchangeable robots}, we mean that the goals for these robots are not fixed {\em a priori}. Instead, all that is required is that each goal assigned to the group is occupied by some robot from the group in the end. An intuitive way to think about this is to view the two groups of robots as two teams with one red team and one blue team. The task is to move the each team of robots from some initial formation to some target formation. Note that if there is a single group of robots, time optimal solution can be computed in polynomial time \cite{YuLav13STAR}. However, once we go from a single group to two groups, this is no longer the case. ~\rqed 

\begin{theorem}\label{t:2g}$\tmpp$ and $\mmpp$ remain NP-complete when robots are partitioned into two or more groups in which robots within each group are interchangeable.
\end{theorem}
{\sc Proof.} In the reduction from $\sat$, let the variable robots belong to one group and the clause robots belong to another group. Note that this does not change the shortest path for any of the robot. To see that this is the case, we first note that each variable robot still requires at least $m + 2$ time steps to go from the left end of a horizontal strip to the right end of a horizontal strip. For a clause robot, although some robots can now reach a goal faster ({\em e.g.}, the distance between $v_{c_1}^g$ and $v_{c_m}$ is only $3$ instead of $m + 2 = 5$), occupying the goal vertex $v_{c_1}^g$ requires at least $m + 2$ time steps. Moreover, this is only possible for the clause robot starting at $v_{c_1}$ to do so. After this, $v_{c_2}^g$ can only be reached by the clause robot at $v_{c_2}$ in $m + 2$ steps. Inductively, a minimum time solution (makespan or total arrival time) requires the clause robot starting at $v_{c_j}$ to go to $v_{c_i}^g$, which requires $m + 2$ time steps.

Through this analysis, we have established the equivalence between many robots and two groups of robots in terms of the reduction from $\sat$ to $\tmpp$ and $\mmpp$, which then shows that $\tmpp$ and $\mmpp$ are NP-complete even when there are only two groups of robots. It is straightforward to add additional robot groups. For example, to have three groups, we may simply split the group of variables robots into two arbitrary (non-empty) groups.~\qed

\remark In viewing the result from \cite{YuLav13STAR} and Theorem~\ref{t:2g}, $\tmpp$ and $\mmpp$ experience a {\em sharp} transition in computational complexity as the robots go from a single group to two groups. The intuitive explanation behind the sudden change is the following. When there is a single group of robots, no two robots will ever need to run in opposite directions. That is, the situation illustrated in Fig.~\ref{fig:moves}(b) never happens when there is a single group of robots ({\em i.e.} all robots are interchangeable), because two such robots can simply ``exchange'' goals and reduce travel distance (and time). The situation in Fig.~\ref{fig:moves}(c) may still happen for a single group of robots, but it is possible to show that there is only a limited number (no more than the total number of robots) of such ``meet'' conflicts. Moreover, these conflicts can be resolved globally to obtain an optimal solution.~\rqed

\section{Computational Complexity of Distance-Optimal $\mpp$ Formulations}\label{sec:distance}

Since computing time-optimal solutions for $\mpp$ is NP-hard, conceivably, computing distance-optimal solutions for $\mpp$ is likely to be an intractable task as well. However, showing this rigorously turns out to be much more challenging. Intuitively, proving that finding distance-optimal solutions is NP-hard is more involved because there is no longer the need to synchronize the movements between all robots time-wise; robots now only affect each other through physical interactions. As a consequence, simulating a circuit like structure, as we have done in the time-optimal case, becomes tricky for the distance-optimal case. To accomplish this, in \cite{YuLav13AAAI}, we have resorted to adopt a complex reduction from $\satt$\footnote{$\satt$ is a specialized, NP-hard version of the boolean satisfiability problem. In a $\satt$, there are $n$ variables and $n$ clauses. Each variable appears exactly four times as literals, twice negated and twice non-negated. Each clause contains exactly four literals.} \cite{RatWar90}. Although we are able to prove the NP-hardness of $\dmpp$ in \cite{YuLav13AAAI}, much of the effort there went to coming up with a  {\em locking mechanism} for preserving the structure of $\satt$ during its reduction to $\dmpp$. The complicated locking mechanism then unfortunately masks the intrinsic structure that makes $\dmpp$ intractable. Moreover, the same structure does not readily extend to show the intractability of $\mmdmpp$. Here, we address these issues with a direct reduction from $\sat$ to $\dmpp$. The update reduction scheme can also be easily adopted to show the NP-hardness of $\mmdmpp$. 

\subsection{Total Distance Objective}
As we want to prove the intractability $\dmpp$ using a reduction from $\sat$, it is essential to preserve the structure that renders $\sat$ hard. Because distance-optimal solutions for $\mpp$ do not require time synchronization, to retain the structure of $\sat$, some form of locking mechanism is necessary to restrict undesirable (time-parametrized) robot paths. We achieve this through the addition of a large number of  robots int the reduced $\mpp$ instance. 

\begin{figure}[htp]
\begin{center}
\includegraphics[width=0.48\textwidth]{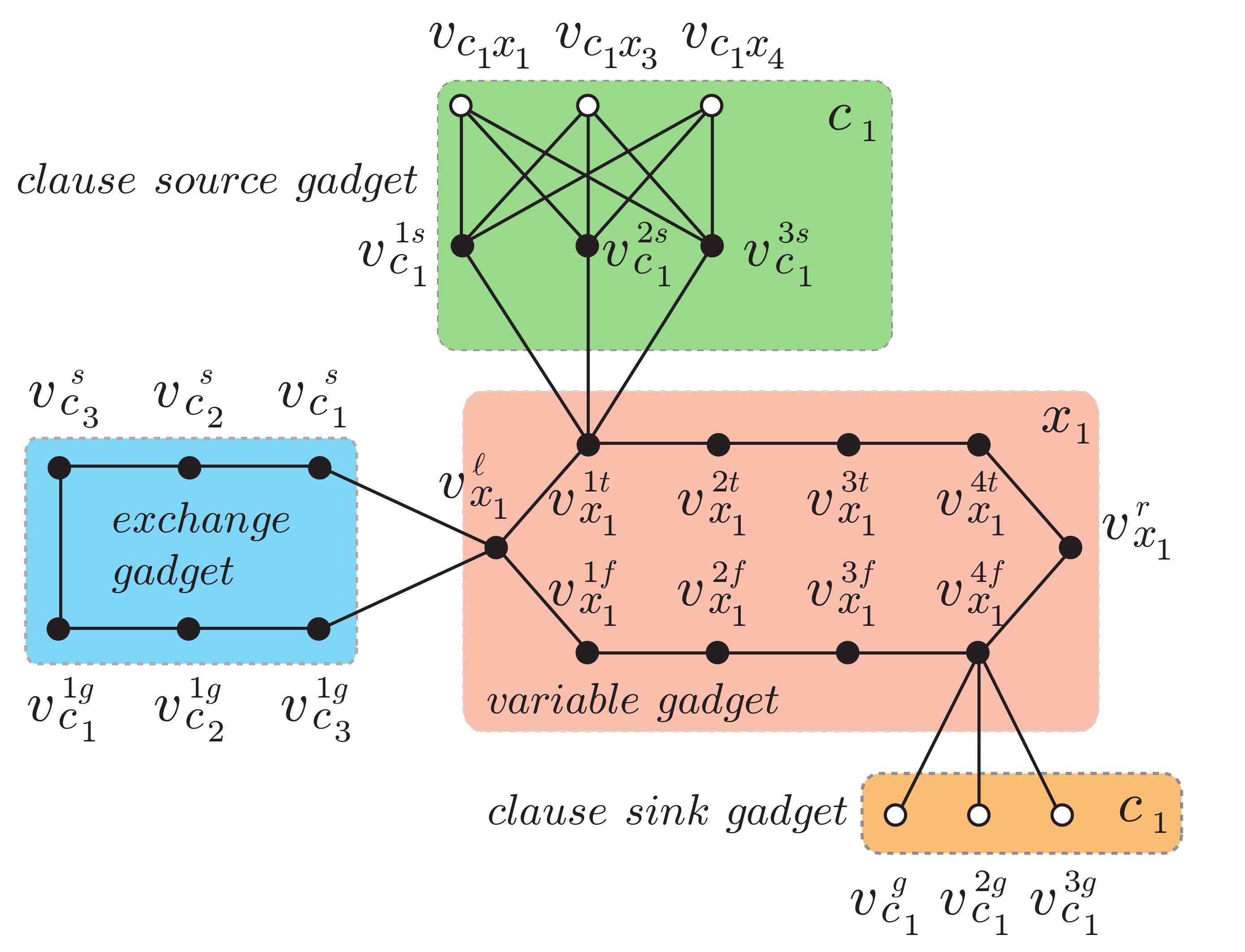} 
\end{center}
\caption{\label{fig:3sat-dmpp} Essential gadgets and their interconnection for the construction of a $\dmpp$ instance from a $\sat$ instance. }
\end{figure}

Fig.~\ref{fig:3sat-dmpp} provides the essential pieces (gadgets) needed for the construction of the $\dmpp$ instance and illustrates how the pieces connect to each other. There are four types of gadgets in the construction: {\em (i)} $n$ {\em variable gadgets}, one of which is shown in the pink block in Fig.~\ref{fig:3sat-dmpp}, {\em (ii)} $m$ {\em clause source gadgets}, one of which is shown in (the green block), {\em (iii)} $m$ {\em clause sink gadgets}, one of which is shown in (the orange block), and {\em (iv)} a single {\em exchange gadget} (the blue block). All vertices in Fig.~\ref{fig:3sat-dmpp} marked with solid circles have robots occupied on them in the beginning. In the following, the construction of each type of gadgets is described in detail. 

{\bf Clause gadgets}. A clause gadget (in the case of Fig.~\ref{fig:3sat-dmpp}, for the clause $c_1$ from~\eqref{equation:sat}) has a source part and a sink part. In general, for a clause $c_i$, the clause sink gadget contains three vertices that are unoccupied in the begining, $v_{c_i}^{g}, v_{c_i}^{2g}$, and $v_{c_i}^{3g}$. The clause source gadget has 6 vertices, partitioned here into two layers of three vertices each. The bottom layer has three vertices named $v_{c_j}^{ks}, 1 \le k \le 3$. The top layer has names depending the actual content of the clause. We use clause $c_1$ from~\eqref{equation:sat} as an example to illustrate the naming scheme. For $c_1 = x_1\vee \neg x_3 \vee x_4$, the top three vertices are named $v_{c_1x_1}$, $v_{c_1x_3}$, and $v_{c_1x_4}$, respectively. There are three robots starting from a clause source gadget, initially residing on vertices $v_{c_j}^{ks}, 1 \le k \le 3$. 

{\bf Variable gadgets.} A variable gadget has the same graph structure, a $(2m + 4)$-cycle, as that from Fig.~\ref{fig:3sat-mpp}. The difference here is that all vertices have robots on them initially; the construction for time-optimal $\mpp$ only has one robot on each of such gadgets. We name the vertices on the $(2m + 4)$-cycle as illustrated in Fig.~\ref{fig:3sat-dmpp}; the superscript letters $r, \ell, t,$ and $f$ represent {\em right}, {\em left}, {\em true}, and {\em false}, respectively. The connections between the variable gadget and the clause source/sink gadgets are similar to that from Fig.~\ref{fig:3sat-mpp} as well, although slightly more complex. Using clause $c_1$ as an example, $v_{c_1}^{1s}, v_{c_1}^{2s}$, and $v_{c_1}^{3s}$ are each connected to $v_{x_1}^{1t}$, $v_{x_3}^{1f}$, and $v_{x_4}^{1t}$. Then $v_{c_1}^{g}$, $v_{c_1}^{2g}$, and $v_{c_1}^{3g}$ are each connected to $v_{x_1}^{4f}$, $v_{x_3}^{4t}$, and $v_{x_4}^{4f}$. Note that vertices $v_{x_3}^{1f}$, $v_{x_4}^{1t}$, $v_{x_3}^{4t}$, and $v_{x_4}^{4f}$, which are on variable gadgets for $x_3$ or $x_4$, are not shown in Fig.~\ref{fig:3sat-dmpp}. 

{\bf Exchange gadget.} The last essential piece of the graph structure, the exchange gadget, has $2m$ vertices, $v_{c_j}^{1s}$ and $v_{c_j}^{1g}$ for $1 \le j \le m$. The vertices $v_{c_1}^{s}$ and $v_{c_m}^{1g}$ are connected to $v_{x_i}^{l}$ for all $1 \le i \le n$. All $2m$ vertices are occupied by robots. 

\begin{figure}[htp]
\begin{center}
\includegraphics[width=0.48\textwidth]{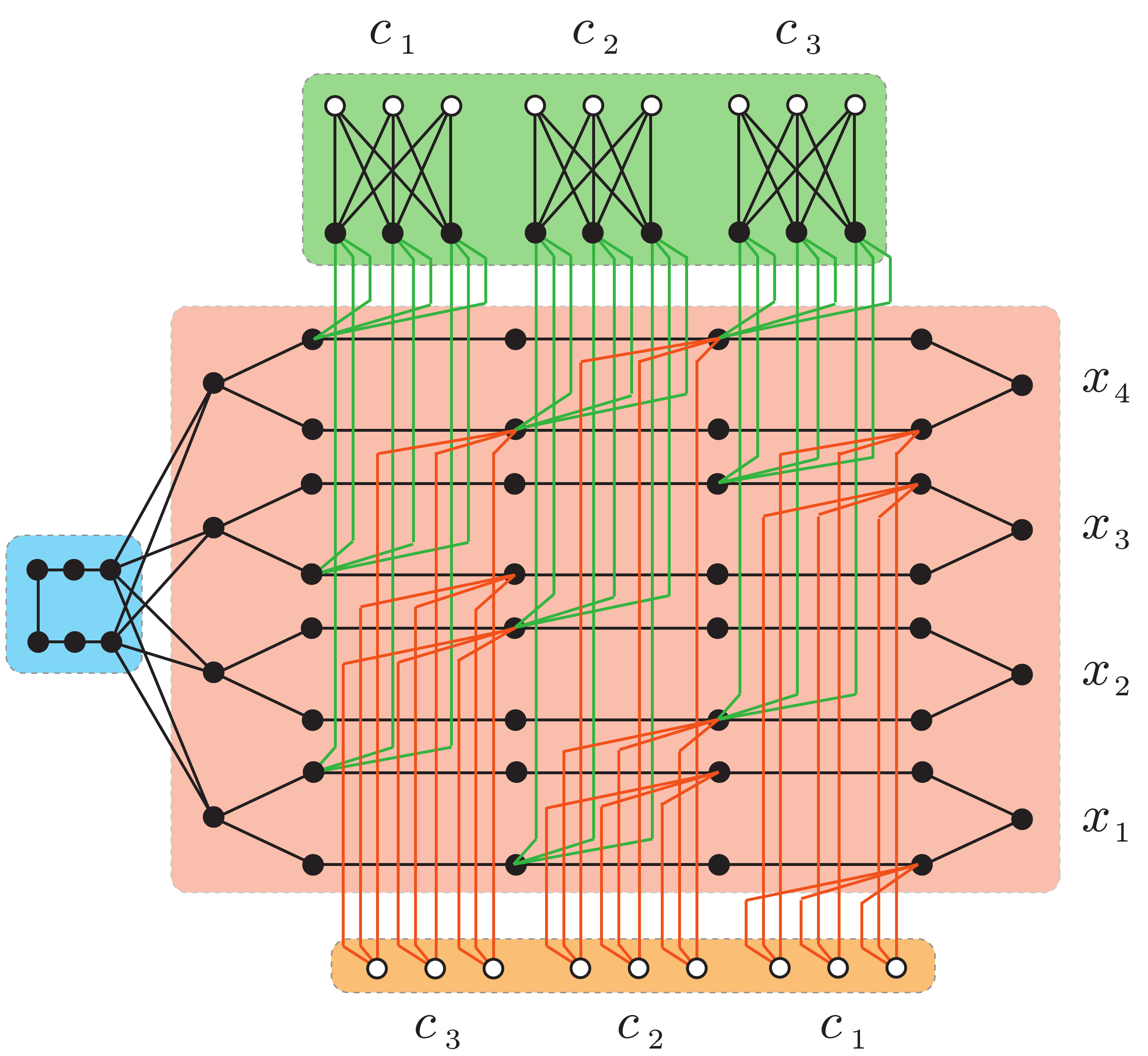} 
\end{center}
\caption{\label{fig:3sat-dmpp-full} The full graph structure and the starting configuration of a $\dmpp$ instance reduced from the $\sat$ instance given in~\eqref{equation:sat}.}
\end{figure}

With the individual gadgets fully specified, we piece together the gadgets for a reduction from $\sat$ to $\dmpp$. 
\begin{theorem}\label{t:dmpp}$\dmpp$ is NP-complete. \end{theorem}
{\sc Proof.} {\bf Instance construction}. The full graph structure of the reduced $\dmpp$ instance from the $\sat$ instance~\eqref{equation:sat}, along with the starting robot locations, is given in Fig.~\ref{fig:3sat-dmpp-full}. To complete the $\dmpp$ instance, we need to specify goals for all robots and then set the value of $K$. 

The goals for the robots are assigned as follows. 
\begin{itemize}
\item[(i)] If a robot start from a vertex named with ``s'' in the superscript, then the goal for the robot is the correspondingly named vertex with a ``g'' in the superscript. For example, the robot starting from $v_{c_1}^{1s}$ must go to $v_{c_1}^{1g}$. 
\item[(ii)] The robots on $v_{c_j}^{1g}$, $1 \le j \le m$ ({\em i.e.}, the vertices on the lower portion of the exchange gadget) must go to $v_{c_j}^{s}$ ({\em i.e.} the diagonal location on the exchange gadget). 
\item[(iii)] For robots on a variable gadget, if a robot resides on a vertex connecting to a clause source gadget, {\em e.g.}, $v_{x_i}^{jt}$ or  $v_{x_i}^{jf}$, the robot then must go to the closest $v_{c_jx_i}$ vertex, which is unique. For example, for the robot starting on $v_{x_1}^{1t}$, its goal vertex is $v_{c_1x_1}$. 
\item[(iv)] For all other robots on a variable gadget, the robot must reach the opposite (farthest) location on the gadget. For example, a robot from $v_{x_1}^{\ell}$ must go to $v_{x_1}^r$. 
\end{itemize}

To set the value of $K$, we collect the minimum distance required for each robot to reach its goal, which is straightforward to observe. Table~\ref{table:distances} lists these distance for most of the robots with the exception of robots starting on a variable gadget. If a robot starts on a variable gadget vertex that is adjacent to a clause source gadget, it only needs to travel two steps to reach its goal ({\em e.g.}, the robot on $v_{x_1}^{1t}$ needs to go $v_{c_1x_1}$). For all other robots on variable gadgets, they need to travel a minimum distance of $m + 2$ to reach the opposite side. The $\dmpp$ instance is fully specified by setting $K$ as the sum of all these minimum individual distances. 

\begin{table}
\begin{center}
	\caption{\label{table:distances}Minimum required distance for robots not starting from a variable gadget.}
	\begin{tabularx}{\columnwidth}{ccXXXX}
   \hline\hline
	 Start vertex  && $v_{c_j}^{1s}$ & $v_{c_j}^{2s}, v_{c_j}^{3s}$ & $v_{c_j}^{s}$ & $v_{c_j}^{1g}$ \\
	 \hline
	 Min distance && $m + 2$ & $m + 4$ & $m + 3$ & $m$ \\
	 \hline\hline
	 \end{tabularx}
\end{center}
\end{table}

{\bf Many-one reduction}. We first show that if the $\sat$ instance has a satisfactory assignment, then all robots can follow their shortest possible paths to reach their respective goals. Since the variable gadgets are filled with robots, for robots to follow shortest possible paths, robots on a single variable gadget can only move in either clockwise or counterclockwise direction but not both (this will be made more formal shortly). The direction with with robots on a variable gadget rotates is decided by the assignment of true or false to the corresponding variable. If a variable is set to true (resp., false), then the direction of rotation on the corresponding variable gadget is counterclockwise (resp., clockwise). Note that this convention is similar to that in the time-optimal case.

Once the direction of rotation along each variable gadget is decided, we use this information to pick the ``correct'' variable gadgets to move the robots starting on $v_{c_j}^{ks}, 1 \le k \le 3$. We point out that the important robots here are the robots starting on $v_{c_j}^{1s}, 1 \le j \le m$ because these robots must end up on the exchange gadget. For such a robot to move to the exchange gadget, it must first reach the left end of a variable gadget. This can be guaranteed as follows. For a clause $c_j$, one of its literal, corresponding to a variable $x_i$, must be true given a satisfiable assignment. We let the robot starting on $v_{c_j}^{1s}$ move to the variable gadget for $x_i$. Now if the literal is $x_i = true$ (resp., $\neg x_i = true$), then the robot starting on $v_{c_j}^{1s}$ will be moved to the upper (resp., lower) path on the $i$-th variable gadget. At the same time, the rotation direction of the variable gadget is counterclockwise (resp., clockwise). Therefore, in either case, the robot starting on $v_{c_j}^{1s}$ can reach the left most vertex ({\em e.g.}, $v_{x_i}^{\ell}$) following the rotation direction of the $i$-th variable gadget. Subsequently, the robot can be moved to the exchange gadget and swap out the robot starting on $v_{c_j}^s$. 

As a concrete example, for $c_1 = x_1\vee \neg x_3 \vee x_4$, if $x_1$ is set to true in a satisfiable assignment, then robot starting at $v_{c_1}^{1s}$ can move to $v_{x_1}^{1t}$ (see Fig.~\ref{fig:3sat-dmpp}), via rotations of robots along the 6-cycle $v_{c_1}^{1s}-v_{x_1}^{1t}-v_{c_1}^{2s}-v_{x_3}^{1f}-v_{c_1}^{3s}-v_{x_4}^{1t}$. The robot can then move to $v_{x_1}^{\ell}$ following the counterclockwise rotation of the variable gadget for $x_1$. 

After describing the key moves, we now provide the full set of movements yielding the minimum total distance solution. First, all robots starting on clause source gadgets are moved out of these gadgets via rotations along $6$-cycles stated above. Then, robots on the variable gadgets will start synchronous rotations. Whenever a robot starting from some $v_{c_j}^{1s}$ reach some $v_{x_i}^{\ell}$, robots on the exchange gadget will rotate counterclockwise to move the robot from $v_{x_i}^{\ell}$ to the exchange gadget. These steps are repeated until the robots starting on $v_{x_i}^{\ell}$ reach $v_{x_i}^{r}$ for all $1 \le i \le n$. Lastly, vertices on the clause sink gadgets get filled with appropriate robots. It is straightforward to check such a plan enables all robots to reach their goals following their respective shortest paths.

Having established the reduction in one direction, we also need to show that if the $\mpp$ instance has a minimum possible total distance solution, then the corresponding $\sat$ instance is satisfiable. To show this, we first establish what possible moves can happen at a given stage if the minimum possible total distance is to be reached. In the beginning, it is clear that robots on the exchange gadget cannot move. Also, rotations along any variable gadget cannot happen either because this will cause robots on vertices like $v_{x_1}^{1t}$ to travel extra distances. The only possible move in the beginning then must involve robots on a clause gadget. Assume the clause source gadget for some $c_j$ is involved. To avoid traveling extra distance, this forces all robots on the clause source gadget $c_j$ to move synchronously. To see this, note that a robot starting from a clause source gadget must move out of the gadget in one step. And this will cause an adjacent robot on a variable gadget to move to the clause source gadget, also due to the minimum distance constraint. This in turn forces another robot on the same clause source gadget to move to a variable gadget. In the case of $c_1$, this means the first move must be a synchronous rotation along a 6-cycle, for example $v_{c_1}^{1s}-v_{x_1}^{1t}-v_{c_1}^{2s}-v_{x_3}^{1f}-v_{c_1}^{3s}-v_{x_4}^{1t}$. 

Once a clause source gadget interacts with (three) variable gadgets, there cannot be further interactions between the clause source gadget and the rest of the $\mpp$ instance graph. After this first step, some variable gadget may start rotating. Before a variable gadget can start rotating, all its interaction with clause source gadgets must be fully complete (because, for example, the robot starting at $v_{x_1}^{1t}$ must be moved out of the variable gadget to avoid traveling extra distances). This also means that there is no interaction ({\em i.e.}, flow of robots) between any two variable gadgets. Therefore, once a variable gadget starts rotating, it only has limited interaction with the exchange gadget until it completes $m + 2$ rotations, all in the same direction (either clockwise or counterclockwise), after which some robots will move out of it to clause sink gadgets. 

To summarize the previous two paragraphs, a clause source gadget can interact (only once and simultaneously) with three variable gadgets in a single synchronized rotation. On the other hand, each variable gadget must first complete its interaction with all (neighboring) clause source gadgets before it can rotate. Then, the variable gadget must decide a direction to rotate for $m + 2$ times. Since we assume that we have a total distance optimal solution, this means the rotation directions of the variable gadgets are known. This in turn means that we know robots on $v_{c_j}^{1s}, 1 \le j \le m$ should go to which variable gadgets so the rotation of the variable gadgets will allow them to travel to their goals on the exchange gadget along a minimum distance path. As we have established in the proof of Theorem~\ref{t:mpp-tt-np-hard}, this is equivalent to finding a satisfactory assignment for the corresponding $\sat$ instance. 

So far we have established the NP-hardness of $\dmpp$. Because $\dmpp$ is easily seen to be in NP, it is NP-complete.~\qed

\subsection{Maximum Distance Objective}
Using the same general technique and with some added effort, we can establish that $\mmdmpp$ is also intractable.

\begin{theorem}\label{t:mdmpp}$\mmdmpp$ is NP-complete. \end{theorem}
{\sc Proof.} The reduction used here is a modification of that from the proof of Theorem~\ref{t:dmpp}. For the given $\sat$ instance~\eqref{equation:sat}, the $\mmdmpp$ instance uses the same four types of gadgets as the $\dmpp$ instance. Moreover, only small parts of the gadgets are changed. In particular, the interconnections between the gadgets ({\em e.g.}, Fig.~\ref{fig:3sat-dmpp-full}) remain identical. 

\begin{figure}[htp]
\begin{center}
\includegraphics[width=0.48\textwidth]{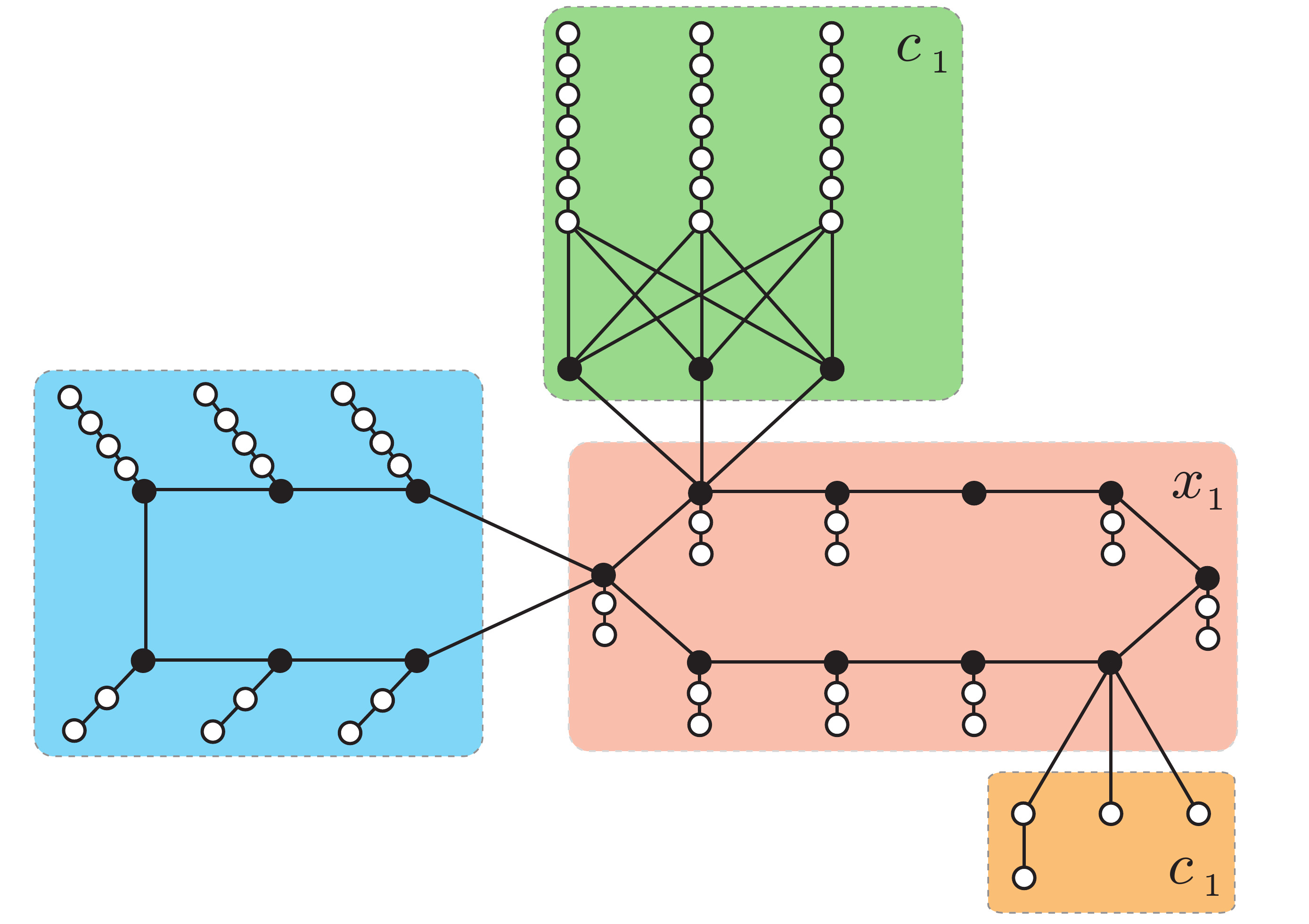} 
\end{center}
\caption{\label{fig:3sat-mmdmpp} Reduction of $\sat$ to $\mmdmpp$ with $K = m + 4$ . }
\end{figure}

The updated construction of the gadgets, adopted from Fig.~\ref{fig:3sat-dmpp}, is given in Fig.~\ref{fig:3sat-mmdmpp}; the vertex labels are omitted. The difference here is that additional one-way paths are appended to some vertices to make all robots travel the same (shortest) distance of $m + 4$. For example, since robots starting from $v_{c_j}^{1s}, 1 \le j \le m$, need to travel $m + 2$ steps to reach their goals in the $\dmpp$ instance, we let these robots travel two additional steps (reflected by the sequence of length two paths in the lower part of the blue exchange gadget). Note that for variable gadgets, we attach a path of length two to each of its vertex unless the vertex is connected to a clause sink gadget. In the case of the variable gadget for $x_1$, no path of length two is added to the third upper vertex ({\em i.e.}, $v_{x_1}^{3t}$ in Fig.~\ref{fig:3sat-dmpp}) and the fourth lower vertex ({\em i.e.}, $v_{x_1}^{4f}$ in Fig.~\ref{fig:3sat-dmpp}). 

We note that each added vertex can only be used for a specific robot. Otherwise, some robot must travel more than a distance of $m + 4$ to reach the designed goal. To complete the proof, we only need to show that if the $\mmdmpp$ instance has a yes answer, then the corresponding $\sat$ instance is satisfiable. We establish this by showing that the newly added vertices and edges do not make the problem easier to solve. The added vertices and edges on a clause source gadget can only be used after the interaction of the clause source with adjacent variable gadgets is complete. Also, the added vertex and edge on a clause sink gadget can only be used after the rotation of a variable gadget is fully complete ({\em i.e.}, $m + 2$ rotations). Therefore, added structures on the clause source/sink gadgets do not allow additional path flexibility. Similarly, the added vertices and edges on the exchange gadget can only be used after all rotations of the exchange gadget are complete ({\em i.e.}, $m$ rotations). Therefore, the added structures on the exchange gadgets do not allow additional path flexibility. 

The case of the variable gadget is more tricky because one variable gadget may finish all its $m + 2$ rotations and then move all the robots out of the $(2m + 4)$-cycle. This potentially can then be used for other robots to pass through. However, we note that any robot that does this must come from a different variable gadget or a non-adjacent clause source gadget. It is straightforward to check that such moves will incur extra travel distance for the robots that are involved. Therefore, the min-max distance requirement does not allow any robot to use an ``un-intended'' variable gadget as part of its path. This then returns us to the $\dmpp$ case. The rest of the proof follows that of Theorem~\ref{t:dmpp}. ~\qed

\section{Conclusion}\label{sec:conclusion}
In this paper, we have investigated the structural and complexity of optimal $\mpp$ problems that minimize the total arrival time, the makespan, the total distance, and the maximum distance for $\mpp$. We show that each pair of these objectives cannot be simultaneously optimized. We then further establish that optimizing over each of these objectives is NP-hard. We conclude that these common time- and distance-based optimality objectives are computationally intractable, suggesting that algorithm designers should look for approximations and heuristics in resolving these problems. 

Our study also raises interesting questions with practical implications; we mention two here. On the structure side, although the objectives are in a sense ``incompatible'', they are nevertheless highly similar. For example, our observation seems to suggest that the time horizon required for minimizing the maximum distance is no more than twice of that needed for minimizing the makespan. A more careful look at such issues could reveal additional structures which may lead to better algorithm design. On the complexity side, unlike the time-optimal case, the NP-hardness proofs for $\dmpp$ and $\mmdmpp$ do not readily extend to show that these problems remain hard for two groups of robots. We believe that this is indeed the case and leave it as a conjecture. 
\begin{conjecture}$\dmpp$ and $\mmdmpp$ remain NP-hard when there are only two groups of robots. 
\end{conjecture}


\bibliographystyle{IEEEtranN}
\bibliography{jingjin}

\end{document}